\documentclass[twoside,journal]{IEEEtran}
\IEEEoverridecommandlockouts

\usepackage[T1]{fontenc}
\usepackage{lmodern}
\usepackage{times}

\usepackage{amsmath}
\usepackage{amssymb}
\usepackage{amsfonts}
\usepackage{amsthm}
\usepackage{mathtools}
\usepackage{bm}
\usepackage{dsfont}
\usepackage{latexsym}
\usepackage{nicefrac}
\usepackage{cases,subeqnarray}
\usepackage{extarrows}

\usepackage{graphicx}
\usepackage{epsfig}
\usepackage{epstopdf}
\usepackage{float}
\usepackage{subfig}
\usepackage{subcaption}
\usepackage{stfloats}
\usepackage{multirow}
\usepackage{bigstrut}
\usepackage{makecell}
\usepackage{array}
\usepackage{booktabs}

\usepackage{color}
\usepackage{colortbl}
\usepackage{textcomp}
\usepackage{pifont}
\usepackage{microtype}
\usepackage{setspace}

\usepackage[noadjust]{cite}
\usepackage{url}
\usepackage{xr-hyper}
\usepackage[pagebackref=false,breaklinks=true,colorlinks,citecolor=blue,linkcolor=blue,bookmarks=true]{hyperref}

\usepackage{multicol}
\usepackage{balance}
\usepackage{flushend}
\usepackage[nameinlink]{cleveref}
\usepackage{etoc}
\usepackage[ruled,vlined]{algorithm2e}
\usepackage{titletoc}
\usepackage{pdfpages}
\usepackage[nointegrals]{wasysym}
\usepackage{catchfilebetweentags}

\usepackage{tikz}
\definecolor{lime}{HTML}{A6CE39}
\DeclareRobustCommand{\orcidicon}{%
\begin{tikzpicture}
\draw[lime, fill=lime] (0,0) 
circle [radius=0.16] 
node[white] {{\fontfamily{qag}\selectfont \tiny ID}};\draw[white, fill=white] (-0.0625,0.095) 
circle [radius=0.007];\end{tikzpicture}
\hspace{-2mm}}
\definecolor{ylp_color1}{RGB}{255,193,193}
\definecolor{ylp_color2}{RGB}{255,228,225}
\usepackage{tcolorbox}
\newtcbox{\mybox}[1][red]{on line, colback = {RGB}{255,228,225}, colframe = {RGB}{255,193,193},  arc=1mm, auto outer arc, boxrule=0.5pt,}

\foreach \x in {A, ..., Z}{%
\expandafter\xdef\csname orcid\x\endcsname{\noexpand\href{https://orcid.org/\csname orcidauthor\x\endcsname}{\noexpand\orcidicon}}
}

\hypersetup{
 colorlinks=true,
 linkcolor=black,
 filecolor=black,
 urlcolor=black,
 citecolor=blue,
}

\theoremstyle{plain}
\newtheorem{assumption}{Assumption}
\newtheorem{theorem}{Theorem}
\newtheorem{thm}{Theorem}[section]
\newtheorem{lemma}{Lemma}
\newtheorem{lemm}{Lemma}
\newtheorem{proposition}{Proposition}
\newtheorem{prop}{Proposition}
\newtheorem{corollary}{Corollary}
\newtheorem{corr}{Corollary}
\newtheorem{definition}{Definition}

\theoremstyle{remark}
\newtheorem{remark}{Remark}
\newtheorem{rem}{Remark}

\crefname{section}{Sec.}{Secs.}
\crefname{subsection}{Sec.}{Secs.}
\crefname{subsubsection}{Sec.}{Secs.}
\crefname{figure}{Fig.}{Figs.}
\crefname{table}{Table}{Tables}
\crefname{equation}{Eq.}{Eqs.}
\crefname{algorithm}{Alg.}{Algs.}
\crefname{algocf}{Alg.}{Algs.}
\crefname{assumption}{Assumption}{Assumptions}
\crefname{theorem}{Theorem}{Theorems}
\crefname{lemma}{Lemma}{Lemmas}
\crefname{proposition}{Proposition}{Propositions}
\crefname{corollary}{Corollary}{Corollaries}
\crefname{remark}{Remark}{Remarks}
\crefname{definition}{Definition}{Definitions}
\crefformat{equation}{#2Eq.~#1#3}
\crefmultiformat{equation}{#2Eqs.~#1#3}{ and~#2#1#3}{, #2#1#3}{ and~#2#1#3}
\Crefformat{equation}{#2Eq.~#1#3}
\Crefmultiformat{equation}{#2Eqs.~#1#3}{ and~#2#1#3}{, #2#1#3}{ and~#2#1#3}

\definecolor{lightkeycolor}{RGB}{255,240,240}
\definecolor{lightgray}{RGB}{234,234,234}

\PassOptionsToPackage{table}{xcolor}
\def\RevisionPreambleLight{1}
\ifdefined\pdfminorversion\pdfminorversion=4\fi
\ifdefined\pdfobjcompresslevel\pdfobjcompresslevel=0\fi

\makeatletter
\PassOptionsToPackage{table}{xcolor}
\@ifpackageloaded{xcolor}{}{\usepackage{xcolor}}
\@ifpackageloaded{soul}{}{\usepackage{soul}}
\@ifpackageloaded{pifont}{}{\usepackage{pifont}}
\@ifpackageloaded{tikz}{}{\usepackage{tikz}}
\makeatother
\definecolor{revisionHighlightColor}{RGB}{0,0,255}
\providecommand{\blue}[1]{}
\renewcommand{\blue}[1]{\textcolor{revisionHighlightColor}{#1}}
\providecommand{\RegisterRevisionHighlightCommands}{%
  \ifdefined\cite\soulregister\cite7\fi
  \ifdefined\ref\soulregister\ref7\fi
  \ifdefined\cref\soulregister\cref7\fi
  \ifdefined\Cref\soulregister\Cref7\fi
  \ifdefined\eqref\soulregister\eqref7\fi
  \ifdefined\textbf\soulregister\textbf7\fi
  \ifdefined\textit\soulregister\textit7\fi
  \ifdefined\emph\soulregister\emph7\fi
}
\RegisterRevisionHighlightCommands
\makeatletter
\@ifundefined{hlmaskenv}{%
  \newsavebox{\hlmasksavebox}%
}{}
\makeatother
\ifdefined\RevisionPreambleLight
  \def\RevisionPreambleMaybeEnd{ }
\else
  \let\RevisionPreambleMaybeEnd\relax
\fi
\RevisionPreambleMaybeEnd

\usepackage{mathtools,bm,amsmath,amssymb,amsfonts,amsthm,nicefrac,latexsym,dsfont,extarrows,bbm}
\usepackage{cases,subeqnarray}
\usepackage{algorithm,algorithmicx,algpseudocode}
\usepackage{booktabs,multirow,subcaption,graphicx,epsfig,epstopdf,float,stfloats,bigstrut,makecell,array}
\usepackage[most]{tcolorbox}
\usepackage{xcolor,color,colortbl,textcomp,pifont,microtype}
\usepackage{multicol,balance,flushend,setspace,titletoc}
\usepackage[noadjust]{cite}
\usepackage{url,xr-hyper}
\makeatletter
\@ifpackageloaded{wasysym}{}{\usepackage[nointegrals]{wasysym}}
\makeatother

\usepackage[pagebackref=false,breaklinks=true,colorlinks=false,pdfborder={0 0 0},bookmarks=true,linkbordercolor=white,citebordercolor=white,urlbordercolor=white,filebordercolor=white]{hyperref}
\makeatletter
\@ifundefined{Hy@setpdfversionfalse}{}{\Hy@setpdfversionfalse}
\makeatother
\usepackage[nameinlink]{cleveref}
\RegisterRevisionHighlightCommands

\definecolor{lime}{HTML}{A6CE39}
\DeclareRobustCommand{\orcidicon}{\begin{tikzpicture}\draw[lime, fill=lime] (0,0) circle [radius=0.16] node[white] {{\fontfamily{qag}\selectfont \tiny ID}};\draw[white, fill=white] (-0.0625,0.095) circle [radius=0.007];\end{tikzpicture}\hspace{-2mm}}
\foreach \x in {A, ..., Z}{\expandafter\xdef\csname orcid\x\endcsname{\noexpand\href{https://orcid.org/\csname orcidauthor\x\endcsname}{\noexpand\orcidicon}}}

\theoremstyle{plain}

\newtheorem{theorem}{Theorem}

\theoremstyle{remark}

\crefname{section}{Sec.}{Secs.}
\crefname{subsection}{Sec.}{Secs.}
\crefname{subsubsection}{Sec.}{Secs.}
\crefname{figure}{Fig.}{Figs.}
\crefname{table}{Table}{Tables}
\crefname{equation}{Eq.}{Eqs.}
\crefname{algorithm}{Alg.}{Algs.}
\crefname{assumption}{Assumption}{Assumptions}
\crefname{theorem}{Theorem}{Theorems}
\crefname{lemma}{Lemma}{Lemmas}
\crefname{proposition}{Proposition}{Propositions}
\crefname{corollary}{Corollary}{Corollaries}
\crefname{definition}{Definition}{Definitions}
\crefformat{equation}{#2Eq.~#1#3}
\crefmultiformat{equation}{#2Eqs.~#1#3}{ and~#2#1#3}{, #2#1#3}{ and~#2#1#3}
\Crefformat{equation}{#2Eq.~#1#3}
\Crefmultiformat{equation}{#2Eqs.~#1#3}{ and~#2#1#3}{, #2#1#3}{ and~#2#1#3}

\definecolor{lightkeycolor}{RGB}{255,240,240}
\definecolor{lightgray}{RGB}{234,234,234}

\def\x{{\mathbf x}}

\ifdefined\pdfminorversion\pdfminorversion=4\fi
\ifdefined\pdfinclusioncopyfonts\pdfinclusioncopyfonts=1\fi
\IfFileExists{macros.tex}{\input{macros}}{}

\let\RevisionPreambleLight\undefined
\IfFileExists{supp.aux}{\externaldocument[supp:]{supp}}{}

\def\x{{\mathbf x}}

\newcounter{tablerow}

\newcommand{\nextrownum}{\stepcounter{tablerow}\thetablerow}
\newcolumntype{N}{>{\nextrownum\hspace{0.5em}}c<{\hspace{0.5em}}}

\begin{document}

\title{PFAdapter: Hierarchical LoRA Decomposition for Personalized Federated MLLMs}

\author{
Jing Liu\orcidA{},~\IEEEmembership{Member,~IEEE},
Kun Yang\orcidB{},
Yan Wang\orcidC{},
Dingkang Yang\orcidD{},
Xiaoshuai Hao\orcidF{},\\
Wei Zhang\orcidE{},
Yang Liu\orcidG{},~\IEEEmembership{Member,~IEEE},~and
Wei Zhou\orcidH{}~\IEEEmembership{Senior Member,~IEEE}
~\thanks{This work was supported by the National Key Research and Development Program of China (Grant No. 2023YFC3604802), National Natural Science Foundation of China (Grant No. 62406075), the Shanghai Key Technology R\&D Program (Grant No. 25511107200), and the Mitacs Elevete, Canada (Grant No. IT44479). (Corresponding author: Yang Liu.)}
\IEEEcompsocitemizethanks{
\IEEEcompsocthanksitem Jing Liu is with the College of Future Information Technology, Fudan University, Shanghai 200433, China, also with the Division of Natural and Applied Sciences, Duke Kunshan University, Suzhou 215316, China, and also with the Department of Electrical and Computer Engineering, The University of British Columbia, BC V6T 1Z4, Canada (e-mail: jing.liu@ieee.org).
\IEEEcompsocthanksitem Kun Yang is with the Ant Group, also with the College of Information Science and Electronic Engineering, Zhejiang University, Hangzhou 310013, China (e-mail: kunyang20@zju.edu.cn).
\IEEEcompsocthanksitem Yan Wang is with the School of Data Science and Engineering, East China Normal University, Shanghai 200062, China (e-mail: yanwang@dase.ecnu.edu.cn).
\IEEEcompsocthanksitem Dingkang Yang is with the College of Intelligent Robotics and Advanced Manufacturing, Fudan University \& Fysics AI, Shanghai 200433, China (e-mail: dkyang20@fudan.edu.cn).
\IEEEcompsocthanksitem Xiaoshuai Hao is with Xiaomi EV, Xiaomi Campus, Anningzhuang Road, Haidian District, 100085, Beijing, China (e-mail: haoxiaoshuai@xiaomi.com).
\IEEEcompsocthanksitem Wei Zhang is with the Information and Communications Technology Cluster, Singapore Institute of Technology (SIT), Singapore 828608 (e-mail: wei.zhang@singaporetech.edu.sg).
\IEEEcompsocthanksitem Yang Liu is with the College of Electronic and Information Engineering, Tongji University, Shanghai 201804, China (e-mail: yang\_liu@ieee.org).
\IEEEcompsocthanksitem Wei Zhou is with the School of Computational and Mathematical Sciences, Cardiff University, CF24 4AG Cardiff, U.K. (e-mail: zhouw26@cardiff.ac.uk).
}
}

\markboth{IEEE TRANSACTIONS ON COGNITIVE COMMUNICATIONS AND NETWORKS}%
{Author \MakeLowercase{\textit{et al}}: PFAdapter: Hierarchical Adapter Decomposition for PFM}

\IEEEtitleabstractindextext{
\begin{abstract}
Agentic AI systems are reshaping communications and networking by deploying autonomous intelligent agents capable of collaborative learning while maintaining data privacy at network edges. Within distributed network environments, Multimodal Large Language Models (MLLMs) serve as cognitive engines for edge devices, yet federated fine-tuning faces substantial challenges in balancing global knowledge aggregation with local adaptation under heterogeneous network conditions. Conventional federated protocols typically rely on uniform parameter aggregation, which conflates domain-invariant features with client-specific nuances, thereby resulting in suboptimal personalization and excessive communication overhead. To address these challenges, we propose \texttt{PFAdapter}, a communication-efficient framework introducing hierarchical LoRA decomposition to explicitly separate adapter parameters into global-shared and local-private components. Query and key projections are assigned to global synchronization for capturing universal multimodal semantics across the network, while value and output projections remain localized for edge-specific adaptation. Additionally, orthogonality regularization based on the Frobenius norm enforces strict separation between these components, preventing redundant feature learning. Selective aggregation protocols synchronize only global-shared components across the federated network, preserving local expertise and reducing communication costs by nearly 50\%.
Experiments on medical VQA and social multimodal benchmarks show that \texttt{PFAdapter} consistently improves over matched federated LoRA baselines while nearly halving synchronized adapter traffic. Empirical results indicate that projection-level decomposition offers a practical path toward communication-efficient agentic MLLM deployment in resource-constrained edge networks.
\end{abstract}

}

\maketitle
\IEEEdisplaynontitleabstractindextext
\IEEEpeerreviewmaketitle

\section{Introduction}
\label{sec:introduction}

\IEEEPARstart{A}{gentic} AI represents a transformative paradigm for modern communication networks, where autonomous agents collaborate to process multimodal information at network edges while keeping raw training data local to each client \cite{baltruvsaitis2018multimodal,radford2021learning}. Multimodal Large Language Models (MLLMs) serve as cognitive cores for intelligent agents deployed across distributed networking systems, enabling sophisticated reasoning over heterogeneous data modalities including text, images, and sensor data \cite{liu2025storageaware,mao2025optimizing,liu2026semantic}. Urgent deployment scenarios emerge in manufacturing anomaly detection requiring rapid inference under constrained budgets, autonomous driving processing multi-sensor data with stringent latency, and medical diagnostics handling privacy-sensitive imaging across distributed facilities \cite{yang2019federated,gao2025learner}, where Federated Learning (FL) enables collaborative model training without centralized data aggregation \cite{mcmahan2017communicationefficient,cheng2025snowball}. Integrating FL with MLLMs becomes critical for realizing scalable data-local agentic systems across next-generation communication networks \cite{wu2025survey,he2025dualcirculation,liu2026enhancing}.
We use this wording deliberately: FL keeps raw samples on device, but it does not by itself guarantee resistance to update inversion, gradient leakage, or membership inference \cite{liu2025differentially,deng2023federated}. The privacy scope of \texttt{PFAdapter} is therefore limited to decentralized training without centralized raw-data pooling, and the stronger attack-model discussion is deferred to \cref{subsec:discussion}.

Deploying federated MLLMs across heterogeneous edge networks introduces substantial challenges arising from extreme variations in local data distributions and network conditions. Medical edge devices often process specialized imaging data that varies significantly across equipment manufacturers and patient demographics \cite{lau2018dataset,liu2021slake,liu2025multimodala}, while IoT nodes in social sensing networks must handle diverse cultural contexts and evolving linguistic patterns \cite{kiela2020hateful,alam2018crisismmd}. Standard global models frequently fail to achieve optimal performance on specialized edge tasks due to uniform aggregation strategies that ignore local data characteristics. Consequently, developing personalized federated learning approaches becomes imperative for edge intelligence systems, enabling adaptation to local nuances while benefiting from collective knowledge \cite{fallah2020personalized,t.dinh2020personalized,zhang2025personalized}.

Fundamental technical barriers hinder effective MLLM personalization within distributed communication networks. Balancing generalizable multimodal representations against edge-specific task expertise poses a primary challenge for network-deployed agents \cite{wilson2020survey,he2025hivefl}, where uniform parameter aggregation induces catastrophic weight washing that dilutes critical client-specific patterns through global averaging \cite{cheng2025federated,jia2025joint}. In manufacturing anomaly detection, extended local training capturing equipment-specific fault signatures paradoxically degrades targeted performance after aggregation while fundamentally undermining model reliability by erasing calibrations essential for safety-critical autonomous systems~\cite{liu2025networking,liu2025anomaly}. Although earlier FL approaches attempted mitigation through proximal regularization \cite{li2020federated}, multi-task learning \cite{zhang2021survey}, and meta-learning \cite{fallah2020personalized}, high-dimensional MLLM parameter spaces frequently lead to insufficient personalization or catastrophic forgetting across the federated network \cite{yao2025federatedllm}.
Communication overhead represents a critical bottleneck for federated MLLM deployment across bandwidth-constrained edge networks. Even when employing Parameter-Efficient Fine-Tuning (PEFT) techniques like Low-Rank Adaptation (LoRA) \cite{hu2021lora}, MLLM backbones impose prohibitive synchronization costs for resource-limited edge devices \cite{liu2025storageaware}. Under typical industrial edge deployments with limited bandwidth (\textit{e.g.}, 5-20 Mbps in manufacturing facilities or vehicular networks), transmitting hundreds of megabytes of adapter parameters per round incurs substantial delays that accumulate across training iterations, proving operationally unacceptable for time-sensitive applications like autonomous vehicle fleets or manufacturing anomaly detection \cite{lim2020federated,liu2024generalized}. Existing solutions frequently overlook redundancy within adapter parameters by transmitting all modules uniformly despite certain components capturing universal features while others remain highly task-specific \cite{fallah2020personalized,wu2025survey}, consequently wasting scarce network resources.

\begin{figure}[t]
    \centering
    \includegraphics[width=\linewidth]{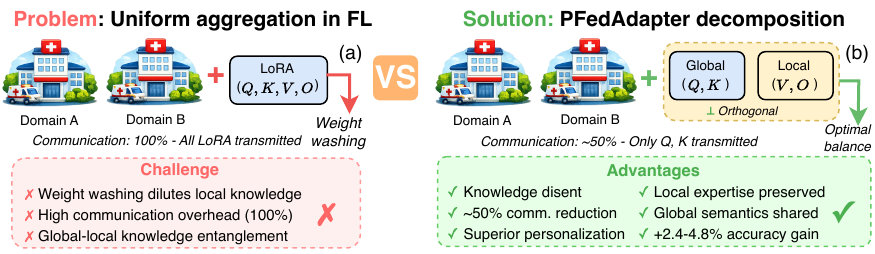}
    \caption{Illustration of challenges in federated MLLM fine-tuning and our proposed solution. (a) Existing FL methods uniformly aggregate all model parameters, leading to high communication costs, loss of client-specific knowledge, and poor personalization under heterogeneous data distributions. (b) \texttt{PFAdapter} introduces hierarchical LoRA decomposition with selective aggregation, where only global-shared adapters are synchronized, while local-private adapters remain on-device.
    }
    \vspace{-2px}
    \label{fig:motivation}
    \vspace{-10px}
\end{figure}

Recent advances in federated MLLM fine-tuning have introduced various strategies to mitigate communication challenges. \texttt{FedAvg} \cite{mcmahan2017communicationefficient} establishes the baseline for collaborative learning across distributed networks, whereas \texttt{FedProx} \cite{li2020federated} incorporates proximal regularization for handling system and statistical heterogeneity. More recently, frameworks including \texttt{FlexLoRA} \cite{bai2024federated} and \texttt{FedMLLM} \cite{xu2025fedmllm} have explored LoRA-based adaptation to reduce trainable parameters. Systematic analysis reveals that uniform aggregation of all LoRA modules conflates domain-invariant features with client-specific nuances \cite{yi2024pfedlora,wang2024flora}. Additionally, methods including \texttt{FedPer} \cite{arivazhagan2019federated} and \texttt{FedRep} \cite{collins2021exploiting} attempt model splitting into global and local layers, yet their application to complex MLLM architectures remains suboptimal. Observed limitations highlight critical gaps in achieving precise knowledge disentanglement within the adapter space.

To address communication and personalization challenges in edge networks, we propose \texttt{PFAdapter}, a resource-efficient federated learning framework for agentic MLLM deployment via hierarchical adapter decomposition. As illustrated in \cref{fig:motivation}, our framework explicitly separates adapter parameters into global-shared and local-private sets based on functional roles of self-attention projections, where query and key projections ($q_p, k_p$) capture universal multimodal semantics for global synchronization while value and output projections ($v_p, o_p$) remain localized for edge-specific adaptation. Orthogonality regularization enforces strict separation by minimizing inner products between global and local parameter matrices, thereby reducing interference in respective feature subspaces while maximizing complementary representation capacity and preventing redundant feature learning.
The main contributions of this work are summarized as follows:
\begin{itemize}
    \item We introduce a novel architectural decomposition for MLLM adapters, categorizing projection modules into global and local components based on their functional roles in capturing multimodal semantics versus edge-specific features.
    \item We propose Frobenius-norm based orthogonality regularization to minimize correlation between global and local adapter weights, ensuring precise knowledge separation for distributed agents.
    \item We develop a communication-efficient synchronization strategy that transmits only global-shared parameters, reducing network traffic by nearly 50\% while preserving local edge expertise.
    \item
    We conduct evaluations across medical VQA and social multimodal benchmarks, showing consistent accuracy, robustness, and communication-efficiency gains under matched federated LoRA protocols.
\end{itemize}

The remainder of this paper is organized as follows. \cref{sec:related_work} reviews related work in federated learning and MLLM fine-tuning. \cref{sec:preliminaries} provides the necessary preliminaries and problem formulation. \cref{sec:method} details the proposed \texttt{PFAdapter} framework and its technical components. \cref{sec:experiments} describes the experimental setup and analyzes the results. Finally, \cref{sec:conclusion} concludes the paper. Additional theoretical analysis, implementation details, and extended experimental results are provided in the supplementary material, \cref{supp:sec:supp_theory,supp:sec:supp_protocol_results,supp:sec:supp_extended_interpretation}.

\section{Related Work}
\label{sec:related_work}

\subsection{Federated learning for large language models}\label{subsec:related_llm}
Federated fine-tuning of Large Language Models (LLMs) and MLLMs has emerged as a critical research direction due to the growing demand for collaborative learning across institutions that cannot pool raw multimodal data centrally \cite{yao2025federatedllm,liu2026collaborative}. Early federated learning frameworks such as \texttt{FedAvg} \cite{mcmahan2017communicationefficient} established the foundation for decentralized model training, yet they were designed for relatively small-scale neural networks and homogeneous data distributions. The emergence of parameter-efficient fine-tuning techniques, particularly LoRA \cite{hu2021lora}, has enabled practical federated adaptation of billion-scale models by reducing the number of trainable parameters from billions to millions. Recent frameworks such as \texttt{FlexLoRA} \cite{bai2024federated} and split-learning approaches \cite{li2025energyefficient} have demonstrated the potential of LoRA-based federated fine-tuning for heterogeneous tasks and resource-constrained environments. However, recent work has identified significant challenges in applying standard FL protocols to MLLMs, including catastrophic forgetting of global knowledge during local updates and the "weight washing" phenomenon where client-specific adaptations are diluted through uniform aggregation \cite{cheng2025federated}. Several approaches have attempted to address these limitations through regularization techniques, multi-stage training protocols \cite{yang2025surveya}, and hybrid architectures that separate feature extractors from task-specific heads. Nevertheless, existing methods treat adapter parameters as a monolithic entity without considering the functional heterogeneity within different projection modules of the self-attention mechanism. In contrast, \texttt{PFAdapter} explicitly recognizes that query and key projections tend to capture structural multimodal relationships that are more amenable to global aggregation, whereas value and output projections are inherently more task-specific and should remain personalized.

\subsection{Personalized federated learning}\label{subsec:related_pfl}
 Personalization in federated learning aims to balance the acquisition of global knowledge with the preservation of local expertise, particularly important for non-IID data distributions \cite{wu2025survey}. Among pioneering approaches, \texttt{FedPer} \cite{arivazhagan2019federated} introduced the concept of maintaining personalized layers at each client while aggregating only the base model parameters. Building upon this foundation, \texttt{FedRep} \cite{collins2021exploiting} proposed a representation learning approach where feature extractors are synchronized globally and classifiers remain local. Recent work has explored more sophisticated personalization strategies including prototype-based learning \cite{tan2022fedproto}, meta-learning frameworks \cite{fallah2020personalized}, and dual-prompt optimization \cite{zhang2025personalized,bai2024federated}. Furthermore, \texttt{FedBABU} \cite{oh2021fedbabu} demonstrated that keeping the model head frozen during server aggregation can significantly improve personalization performance. For multimodal scenarios, \cite{wang2024notall} introduced task-similarity-aware model aggregation for heterogeneous multi-modal clients. However, these methods primarily focus on architectural separation at the layer level rather than parameter-level decomposition within individual modules. Moreover, layer-wise splitting strategies become inefficient for deep transformer architectures where task-specific knowledge is distributed across multiple layers. \texttt{PFAdapter} addresses this limitation by performing fine-grained decomposition at the projection module level, enabling more nuanced control over which aspects of the model are shared versus personalized.
Compared with \texttt{FedPer} and \texttt{FedRep} \cite{arivazhagan2019federated,collins2021exploiting}, which personalize entire layers or heads, \texttt{PFAdapter} moves the personalization boundary inside each attention block and explicitly separates globally aggregated query/key projections from locally retained value/output projections. Accordingly, the projection-level split is tailored to MLLM adapters, where structural cross-modal alignment must be shared across clients but semantic realization remains strongly client-dependent under non-IID data.

\subsection{Parameter-efficient fine-tuning in federated settings}\label{subsec:related_peft} 
PEFT has become indispensable for adapting large-scale models in resource-constrained environments \cite{kim2025missing}. Beyond LoRA, several PEFT variants have been developed including adapter layers \cite{houlsby2019parameter}, prompt tuning \cite{lester2021power}, prefix tuning \cite{li2021prefix}, and Hadamard adapters \cite{chen2023hadamard}. Recent work has explored the integration of these techniques into federated learning frameworks to reduce communication overhead and computational burden \cite{wu2025survey}. Methods such as \texttt{FedAdapter} \cite{yan2024federa} and \texttt{pFedLoRA} \cite{yi2024pfedlora} have demonstrated the effectiveness of adapter-based personalization for language models. Building upon these foundations, \texttt{FloRA} \cite{wang2024flora} introduced heterogeneous low-rank adaptations for federated LLM fine-tuning, while recent work on adaptive LoRA experts and differentially private federated LoRA \cite{liu2025differentially} have further advanced the field. For multimodal applications, approaches like \texttt{FedDLP} \cite{nguyen2025federated} have employed dual adapters with selective pruning to balance local specialization and global knowledge sharing, while visual-language enhancement systems motivate similarly compact adaptation under edge visual workloads \cite{wu2025clipae}. However, these approaches typically apply uniform aggregation to all adapter parameters, failing to distinguish between domain-invariant and domain-specific components. The \texttt{FedMLLM} framework \cite{xu2025fedmllm} represents the state-of-the-art in federated MLLM fine-tuning, yet it relies on full adapter synchronization that conflates global and local knowledge. Recent theoretical analysis has shown that the optimal aggregation strategy should vary across different parameter subsets based on their sensitivity to local data distributions \cite{fallah2020personalized}. Motivated by this insight, \texttt{PFAdapter} introduces selective aggregation that only synchronizes the global-shared adapter components, reducing communication costs while preserving local expertise.
\Cref{supp:tab:supp_pfadapter_method_comparison} provides the full structured comparison of representative personalized FL and federated LoRA methods \cite{yi2024pfedlora,wang2024flora,nguyen2025federated}. In addition, it contrasts \texttt{PFAdapter} with the closely related \texttt{FedMLLM} setting \cite{xu2025fedmllm}.
\section{Preliminaries}
\label{sec:preliminaries}
\subsection{Problem Formulation}
\label{subsec:problem_formulation}
Consider a federated learning system consisting of a central server and a set of $K$ heterogeneous clients, where each client $k \in \{1, \dots, K\}$ possesses a local multimodal dataset $\mathcal{D}_k$. Given the input space $\mathcal{X}$ comprising image-text pairs and the output space $\mathcal{Y}$ representing textual responses, the local dataset is defined as $\mathcal{D}_k = \{(\mathbf{x}_{k,i}, y_{k,i})\}_{i=1}^{n_k}$, where $n_k$ denotes the number of local samples. Let $f_\theta: \mathcal{X} \rightarrow \mathcal{Y}$ represent an MLLM parameterized by $\theta \in \mathbb{R}^d$. The primary objective in personalized federated learning is to find a set of parameters $\{\theta_1, \dots, \theta_K\}$ that minimize the aggregate empirical risk:
\begin{equation}
    \min_{\{\theta_k\}_{k=1}^K} \sum_{k=1}^K \frac{n_k}{N} \mathbb{E}_{(\mathbf{x}, y) \sim \mathcal{D}_k} [\ell(f_{\theta_k}(\mathbf{x}), y)],
    \label{eq:pfl_objective}
\end{equation}
where $N = \sum_k n_k$ is the total number of samples across all clients and $\ell(\cdot, \cdot)$ denotes the cross-entropy loss function. Unlike standard federated learning which seeks a single global model $\theta^*$, personalized approaches allow for client-specific variations to account for statistical heterogeneity \cite{fallah2020personalized,t.dinh2020personalized}.

\subsection{Low-Rank Adaptation}
\label{subsec:lora}
LoRA serves as the foundational parameter-efficient fine-tuning technique for large-scale models. For a pre-trained weight matrix $W_0 \in \mathbb{R}^{d \times k}$, LoRA represents the weight update $\Delta W$ as the product of two low-rank matrices $B \in \mathbb{R}^{d \times r}$ and $A \in \mathbb{R}^{r \times k}$, where the rank $r \ll \min(d, k)$. The forward pass of a modified linear layer is expressed as:
\begin{equation}
    \mathbf{h} = W_0 \mathbf{x} + \Delta W \mathbf{x} = W_0 \mathbf{x} + BA \mathbf{x}.
    \label{eq:lora_forward}
\end{equation}
During the fine-tuning process, $W_0$ remains frozen while only $A$ and $B$ are updated. In the context of MLLMs, LoRA is typically applied to the projection matrices within the self-attention mechanism, specifically the query ($W_q$), key ($W_k$), value ($W_v$), and output ($W_o$) projections \cite{chen2023hadamard}.

\subsection{Federated Optimization}
\label{subsec:federated_optimization}
Standard federated optimization often employs the \texttt{FedAvg} protocol to synchronize model updates across clients. In each communication round $t$, the server selects a subset of clients $\mathcal{S}_t$ and transmits the current global parameters $\theta^t$. Each selected client performs $E$ epochs of local stochastic gradient descent (SGD) to obtain updated parameters $\theta_k^{t+1}$. The server then aggregates these updates using a weighted average:
\begin{equation}
    \theta^{t+1} = \sum_{k \in \mathcal{S}_t} \frac{n_k}{\sum_{j \in \mathcal{S}_t} n_j} \theta_k^{t+1}.
    \label{eq:fedavg_aggregation}
\end{equation}
However, applying this uniform aggregation to all LoRA parameters in MLLMs often leads to the dilution of local task-specific knowledge, necessitating a more granular approach to parameter management \cite{yi2024pfedlora,wang2024flora}.

\section{Method}
\label{sec:method}

\subsection{Method Overview}
\label{subsec:method_overview}
\begin{figure*}[t]
    \centering
    \includegraphics[width=\linewidth]{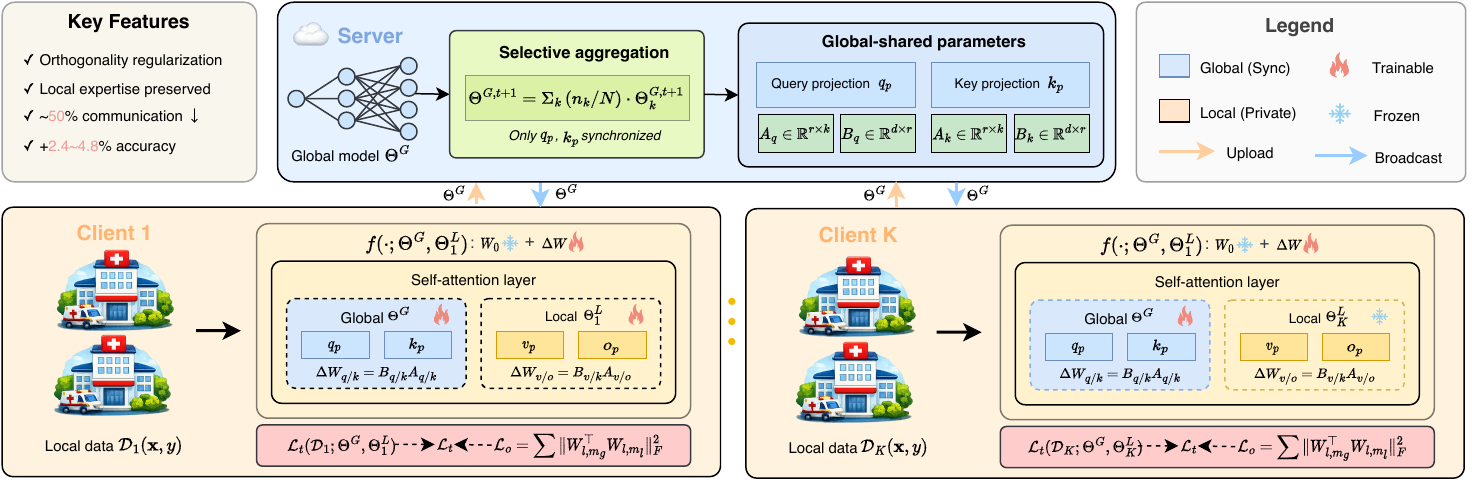}
    \caption{Architecture of \texttt{PFAdapter}. The framework decomposes LoRA adapters into global-shared components ($q_p$, $k_p$) and local-private components ($v_p$, $o_p$). Orthogonality regularization enforces knowledge disentanglement between these sets.
    }
    \vspace{-2px}
    \label{fig:architecture}
    \vspace{-15px}
\end{figure*}

Personalized federated learning for MLLMs requires a delicate balance between global knowledge acquisition and local task adaptation. To achieve this balance, \texttt{PFAdapter} introduces a structured decomposition of the adapter parameter space that explicitly separates domain-invariant features from client-specific nuances. Our design philosophy rests on the observation that different projection modules within the self-attention mechanism exhibit varying degrees of task-specificity. Specifically, the framework partitions the set of LoRA modules into global-shared and local-private subsets based on their functional roles in processing multimodal information. \cref{fig:architecture} illustrates the overall architecture, where query and key projections are synchronized globally while value and output projections are maintained locally. The overall optimization objective for the system is formulated as:
This design differs from layer-wise personalized FL and fully synchronized federated LoRA \cite{arivazhagan2019federated,collins2021exploiting,xu2025fedmllm} because it introduces an intermediate projection-level control point: attention-map formation is shared through query/key adapters, while client-specific representation realization is preserved through value/output adapters. The supplementary structured comparison summarizes this distinction before the formal derivation.
The closest architectural contrast is with \texttt{FedDLP} \cite{nguyen2025federated}: both methods separate shared and personalized adaptation paths, but \texttt{FedDLP} decides what to transmit by pruning an auxiliary shared branch, whereas \texttt{PFAdapter} fixes the communication boundary at the projection type itself and aggregates only the query/key adapters. Consequently, the distinction matters in multimodal non-IID settings because it ties the synchronized subspace directly to attention-relation formation rather than to a sparsified duplicate branch.
\begin{equation}
    \scalebox{0.95}{$\displaystyle\min_{\Theta^G, \{\Theta_k^L\}_{k=1}^K} \sum_{k=1}^K \frac{n_k}{N} \left[ \mathcal{L}_{t}(\mathcal{D}_k; \Theta^G, \Theta_k^L) + \lambda_{o} \mathcal{L}_{o}(\Theta^G, \Theta_k^L) \right]$}
    \label{eq:overall_objective}
\end{equation}
where $\Theta^G$ represents the global-shared parameters and $\Theta_k^L$ denotes the local-private parameters for client $k$. The task loss $\mathcal{L}_{t}$ is typically defined as the negative log-likelihood of the target sequence given the multimodal input:
\begin{equation}
    \scalebox{0.95}{$\displaystyle\mathcal{L}_{t}(\mathcal{D}_k; \Theta) = -\frac{1}{n_k} \sum_{i=1}^{n_k} \sum_{j=1}^{|y_{k,i}|} \log P(y_{k,i,j} \mid y_{k,i,<j}, \mathbf{x}_{k,i}; \Theta)$}
    \label{eq:task_loss_def}
\end{equation}

\subsection{Hierarchical LoRA Decomposition}
\label{subsec:hierarchical_decomposition}
Hierarchical decomposition in \texttt{PFAdapter} formally splits the total set of trainable LoRA parameters $\Theta$ into two disjoint sets: $\Theta^G$ and $\Theta^L$. Let $\mathcal{M} = \{q, k, v, o\}$ denote the set of projection types in the transformer layers. The global parameter set $\Theta^G$ is defined as the union of LoRA weights for query and key projections:
\begin{equation}
    \Theta^G = \bigcup_{l=1}^L \{A_{l,m}, B_{l,m} \mid m \in \{q, k\}\},
    \label{eq:global_params}
\end{equation}
where $L$ is the number of transformer layers. Conversely, the local parameter set $\Theta^L$ contains the weights for value and output projections:
\begin{equation}
    \Theta^L = \bigcup_{l=1}^L \{A_{l,m}, B_{l,m} \mid m \in \{v, o\}\}.
    \label{eq:local_params}
\end{equation}
The intuition behind this specific split arises from the functional roles of self-attention modules. Query and key projections encode the \textit{attention patterns} that determine which tokens attend to each other, capturing structural relationships between multimodal tokens that reflect domain-invariant correspondence patterns \cite{vaswani2017attention,liu2026projecting}. Recent work on adapter decomposition has shown that attention patterns exhibit higher cross-domain transferability than value representations \cite{hu2021lora,houlsby2019parameter}. The attention score matrix $S$ is computed as:
\begin{equation}
    S = \frac{(W_q + \Delta W_q^G)\mathbf{x} ((W_k + \Delta W_k^G)\mathbf{x})^\top}{\sqrt{d_k}}.
    \label{eq:attention_score}
\end{equation}
In contrast, value and output projections encode the \textit{semantic content} that is transformed into the final representation, making them inherently more task-specific and susceptible to local data variations \cite{mickus2024role}. The final attended representation $V'$ is obtained via:
\begin{equation}
    V' = \text{Softmax}(S) (W_v + \Delta W_v^L)\mathbf{x}.
    \label{eq:weighted_sum}
\end{equation}
Consistent with this mechanism, the design aligns with empirical findings that value layers capture more task-specific information than query-key pairs in multi-task learning scenarios \cite{zhang2021survey,koo2025loraa2}.
Under non-IID multimodal federation \cite{li2020federated,hsu2019measuring,liu2026diffusionguided}, the server needs to preserve client-agnostic alignment cues while avoiding over-averaging client-specific semantics. \Cref{eq:attention_score,eq:weighted_sum} make this separation explicit: query/key adapters perturb the attention logits that determine \emph{which} visual-textual tokens interact, whereas value/output adapters govern \emph{what} task-specific content is injected after the shared attention map is formed \cite{vaswani2017attention,mickus2024role}. We therefore assign $\Theta^G$ in \cref{eq:global_params} to the transferable relation-encoding subspace and keep $\Theta^L$ in \cref{eq:local_params} client-resident to absorb label skew, vocabulary bias, and modality imbalance without washing out local semantics during aggregation.
To keep the notation consistent throughout the remainder of the paper, $\Theta^L$ in \cref{eq:local_params} denotes the \emph{structural} set of local-private slots, while the actual parameters owned by client $k$ are written as $\Theta_k^L$. Accordingly, all optimization objectives, algorithmic updates, and convergence statements below use the pair $(\Theta^G,\Theta_k^L)$ when referring to a concrete client state, consistent with personalized FL notation where local client states remain distinct from the shared server-side model~\cite{arivazhagan2019federated,collins2021exploiting}.
For a given input $\mathbf{x}$, the forward pass of a decomposed self-attention layer is expressed as:
\begin{equation}
    \scalebox{0.95}{$\displaystyle\begin{aligned}
        \text{Attn}(\mathbf{x}) &= \text{Softmax}\left(\frac{(W_q + \Delta W_q^G)\mathbf{x} ((W_k + \Delta W_k^G)\mathbf{x})^\top}{\sqrt{d_k}}\right) \\
        &\quad \times (W_v + \Delta W_v^L)\mathbf{x}
    \end{aligned}$}
    \label{eq:decomposed_forward}
\end{equation}
where $\Delta W_m = B_m A_m$ represents the low-rank update for module $m$. Given the resulting partition, the local model for client $k$ is represented as the combination $f(\cdot; \Theta^G, \Theta_k^L)$. During the training process, only $\Theta^G$ is subject to federated aggregation, while $\Theta_k^L$ remains resident on the client device to preserve personalized features. Moreover, the rank $r$ for both global and local adapters is kept consistent to maintain architectural symmetry. Consequently, the total number of trainable parameters per client remains identical to standard LoRA-based federated learning, ensuring no additional memory overhead during local training.

\subsection{Orthogonality-Driven Knowledge Disentanglement}
\label{subsec:orthogonality}
Effective knowledge disentanglement requires that the global and local components learn non-redundant features. To enforce this separation, \texttt{PFAdapter} incorporates an orthogonality regularization term into the local optimization objective. Let $W_{l,m} = B_{l,m} A_{l,m}$ denote the effective weight update for layer $l$ and module $m$. The orthogonality loss $\mathcal{L}_{o}$ is formulated using the Frobenius norm of the product between global and local weight matrices:
\begin{equation}
    \mathcal{L}_{o} = \sum_{l=1}^L \sum_{m_g \in \{q, k\}} \sum_{m_l \in \{v, o\}} \| W_{l,m_g}^\top W_{l,m_l} \|_F^2.
    \label{eq:ortho_loss}
\end{equation}
Minimizing this term encourages the column spaces of global and local adapters to be orthogonal, thereby preventing the local modules from re-learning information already captured by the global components. The Frobenius norm $\| \cdot \|_F$ for a matrix $A \in \mathbb{R}^{m \times n}$ is defined as:
\begin{equation}
    \| A \|_F = \sqrt{\sum_{i=1}^m \sum_{j=1}^n a_{ij}^2}.
    \label{eq:frobenius_norm_def}
\end{equation}
Differentiability of the Frobenius norm allows for efficient gradient-based optimization. During local backpropagation, the gradients of the orthogonality loss with respect to the global and local parameters are computed as:
\begin{equation}
    \nabla_{\Theta^G} \mathcal{L}_{o} = 2 \sum_{l, m_g, m_l} W_{l,m_l} W_{l,m_l}^\top W_{l,m_g},
    \label{eq:grad_global_ortho}
\end{equation}
\begin{equation}
    \nabla_{\Theta^L} \mathcal{L}_{o} = 2 \sum_{l, m_g, m_l} W_{l,m_g} W_{l,m_g}^\top W_{l,m_l}.
    \label{eq:grad_local_ortho}
\end{equation}
Consequently, the total local loss function for client $k$ becomes:
\begin{equation}
    \mathcal{L}_{total} = \mathcal{L}_{t}(\mathcal{D}_k; \Theta^G, \Theta_k^L) + \lambda_{o} \mathcal{L}_{o},
    \label{eq:total_loss}
\end{equation}
where $\lambda_{o}$ is a hyperparameter controlling the strength of the disentanglement constraint. Regularization via orthogonality ensures that the local adapters focus exclusively on domain-specific nuances that cannot be captured by the global model. Moreover, the orthogonality constraint facilitates more stable federated aggregation by reducing the variance of local updates in the global parameter space.
This regularizer also sharpens the global-local interpretation behind the Q/K versus V/O split: when query/key updates already explain a shared attention relation, the penalty discourages value/output adapters from redundantly encoding the same direction, forcing them to capture residual client-specific semantics instead \cite{wu2023orthogonal}. In gradient terms, \cref{eq:grad_global_ortho,eq:grad_local_ortho} project each update away from directions already occupied by its counterpart, thereby reducing the cosine overlap between global-shared and local-private descent steps before aggregation. As a result, the selective aggregation step operates on a subspace whose cross-client bias is controlled, which is the quantity explicitly bounded in our convergence analysis below.

\subsection{Selective Aggregation and Communication Efficiency}
\label{subsec:selective_aggregation}
Selective aggregation protocols in \texttt{PFAdapter} significantly reduce communication overhead while maintaining high performance. In each round $t$, the server only collects and averages the global-shared parameters $\Theta_k^{G, t+1}$ from the participating clients. The aggregation rule is defined as:
\begin{equation}
    \Theta^{G, t+1} = \sum_{k \in \mathcal{S}_t} \frac{n_k}{\sum_{j \in \mathcal{S}_t} n_j} \Theta_k^{G, t+1}.
    \label{eq:selective_aggregation}
\end{equation}
Meanwhile, the local parameters $\Theta_k^L$ are updated locally and never transmitted to the server. Let $C_{total}$ denote the communication cost of standard federated LoRA tuning, where all adapter parameters are synchronized. The communication cost of \texttt{PFAdapter}, denoted as $C_{PFed}$, is given by:
\begin{equation}
    \scalebox{0.94}{$\displaystyle
    \begin{aligned}
        P_{\mathrm{all}} &= \sum_{l=1}^{L} \sum_{m \in \{q,k,v,o\}} r(d_{l,m}^{\mathrm{in}} + d_{l,m}^{\mathrm{out}}), \\
        P_{\mathrm{PF}} &= \sum_{l=1}^{L} \sum_{m \in \{q,k\}} r(d_{l,m}^{\mathrm{in}} + d_{l,m}^{\mathrm{out}}), \\
        C_{PFed}^{(\mathrm{round})} &= 2 b P_{\mathrm{PF}} = \frac{P_{\mathrm{PF}}}{P_{\mathrm{all}}} C_{total}^{(\mathrm{round})} \approx \frac{2}{4} C_{total}^{(\mathrm{round})},
    \end{aligned}$}
    \label{eq:comm_cost}
\end{equation}
Here $r$ is the LoRA rank \cite{hu2021lora}, $d_{l,m}^{\mathrm{in}}$ and $d_{l,m}^{\mathrm{out}}$ are the input/output dimensions of projection $m$ at layer $l$, and $b$ is the transmitted bytes per parameter. Because the four self-attention projections in the deployed MLLM use the same LoRA rank and matched hidden dimensions, synchronizing only $\{q,k\}$ yields the exact parameter-count ratio $P_{\mathrm{PF}}/P_{\mathrm{all}} = 2/4 = 0.5$. The measured traffic in our implementation is therefore reduced from 617 MB/round to 315 MB/round, corresponding to 30.85 GB versus 15.75 GB over 50 rounds, with the small deviation from an ideal 50.0\% explained by serialization and packet rounding overhead.
Because no server-side momentum or adaptive second-moment state is maintained in this branch, \texttt{PFAdapter} uses sample-count-weighted averaging on the uploaded query/key adapters.
Furthermore, the preservation of $\Theta_k^L$ ensures that the model retains its personalized expertise across communication rounds, mitigating the negative effects of weight washing. Given the massive scale of MLLM backbones, such reductions in communication traffic are critical for deployment in resource-constrained environments. Moreover, the selective aggregation strategy prevents the global model from being corrupted by highly specialized local features that do not generalize across the client population.

\subsection{Theoretical Analysis}
\label{subsec:theoretical_analysis}
Convergence analysis of \texttt{PFAdapter} can be established under standard assumptions of smoothness and bounded variance. Let $F(\Theta^G, \{\Theta_k^L\})$ denote the global objective function. Given that the orthogonality regularization is a smooth function of the parameters, the local updates follow a descent direction for the regularized objective. Furthermore, the selective aggregation of $\Theta^G$ can be viewed as a block-coordinate descent step in the parameter space.
We make the non-IID setting explicit through four assumptions: (A1) each local objective $F_k$ is $L_F$-smooth, (A2) stochastic gradients satisfy $\mathbb{E}\|g_k^t-\nabla F_k\|^2 \leq \sigma^2$ and $\|\nabla F_k\| \leq G$, (A3) client heterogeneity is bounded by $\frac{1}{K}\sum_{k=1}^{K}\|\nabla F_k(\Theta^G,\Theta_k^L)-\nabla F(\Theta^G,\{\Theta_j^L\})\|^2 \leq \delta^2$, and (A4) the selective-aggregation bias and orthogonality gradient are bounded as $\|b_t\| \leq \beta$ and $\|\nabla \mathcal{L}_o\| \leq H_o$. Here $\beta$ is treated as an explicit bounded-bias assumption for the projection-restricted server update, not as a theorem already derived from $\delta$, local epochs, or $\lambda_{\text{o}}$. Assumption (A3) does not require IID data, and it only requires the cross-client drift induced by non-IID partitions to remain bounded \cite{li2020federated,hsu2019measuring}.
Theoretical results indicate that the framework achieves a convergence rate of $\mathcal{O}(1/\sqrt{T})$ for non-convex objectives, matching the performance of standard federated learning while providing superior personalization guarantees.

\begin{theorem}[Convergence of \texttt{PFAdapter}]
\label{thm:convergence}
Assume (A1)--(A4) above and choose $\eta_t = c/\sqrt{T}$ with $0 < c \leq 1/(8L_F)$. For the iterates, the averaged stationarity measure satisfies:
\begin{equation}
\begin{aligned}
    \frac{1}{T}\sum_{t=0}^{T-1}
    \mathbb{E}\!\left[\left\| \nabla F(\Theta^{G,t}, \{\Theta_k^{L,t}\}) \right\|^2\right]
    &\leq \frac{C_1 + C_2\delta^2 + C_3\frac{\sigma^2}{|\mathcal{S}_t|}}{\sqrt{T}} \\
    &\quad + C_4\lambda_o^2 H_o^2 + C_5\beta^2,
\end{aligned}
    \label{eq:convergence_rate}
\end{equation}
where 
$C_1 = 2(F^0-F^\star)/c$, $C_2 = 4cL_F$, $C_3 = 4cL_F$, $C_4 = 4(1+cL_F)$, and $C_5 = 4(1+cL_F)$.
Consequently, $\min_{0 \leq t < T}\mathbb{E}[\|\nabla F(\Theta^{G,t},\{\Theta_k^{L,t}\})\|^2] = \mathcal{O}(T^{-1/2})$ whenever the non-IID drift $\delta^2$, orthogonality-gradient magnitude $H_o$, and assumed selective-aggregation bias bound $\beta$ remain finite.
\end{theorem}
A detailed proof sketch, including the descent inequality and telescoping derivation, is provided in the supplementary material, \cref{supp:sec:supp_theory,supp:sec:supp_revision_1}. The supplementary bias appendix also separates a heuristic dependence discussion from the formal theorem assumptions, so the role of $\beta$ is not overstated. Furthermore, the assumptions and theorem statement are placed next to the method definition so that the convergence guarantee remains visible where the selective-aggregation mechanism is introduced, following standard non-IID FL analyses that separate stochastic variance from client-drift terms~\cite{li2020federated,hsu2019measuring}.

\section{Experiments}
\label{sec:experiments}

\subsection{Experimental Setup}
\label{sec:experimental_setup}

\noindent\textbf{Datasets and evaluation protocols.}\label{subsec:datasets} We evaluate \texttt{PFAdapter} on four diverse multimodal benchmarks to assess its effectiveness across medical imaging and social media domains. VQA-RAD \cite{lau2018dataset} comprises 3,515 question-answer pairs across 315 radiology images, emphasizing specialized clinical reasoning and medical domain knowledge. SLAKE \cite{liu2021slake} provides 14,028 samples with 642 images in a bilingual medical VQA setting, offering more complex semantic structures for evaluation. Hateful Memes \cite{kiela2020hateful} contains 10,000 multimodal entries requiring joint text-image processing for hate speech detection in social media contexts. CrisisMMD \cite{alam2018crisismmd} consists of 16,080 image-text pairs from disaster scenarios, categorized into humanitarian assistance tasks including damage severity assessment and resource needs identification.
Performance metrics included Accuracy and weighted F1-score for VQA-RAD, SLAKE, and CrisisMMD, whereas Hateful Memes reported Accuracy together with Area Under the ROC Curve (AUC) instead of F1-score for the binary classification task.
All experiments followed the Aligned modal scenario with Dirichlet concentration parameter $\alpha=0.5$ to simulate moderate data heterogeneity, matching the evaluation protocol established in prior federated MLLM work \cite{xu2025fedmllm}.
Detailed preprocessing, partition reuse, and local-test evaluation protocol notes are provided in the supplementary material, \cref{supp:sec:supp_revision_2}, for VQA-RAD~\cite{lau2018dataset}, SLAKE~\cite{liu2021slake}, Hateful Memes~\cite{kiela2020hateful}, and CrisisMMD~\cite{alam2018crisismmd}.

\noindent\textbf{Baseline methods and implementation configuration.}\label{subsec:baselines}
To benchmark \texttt{PFAdapter} against established federated personalization strategies, we compare against six representative references spanning standard averaging, adaptive server optimization, and personalized LoRA tuning. Zero-shot performance of the pre-trained base model served as the lower bound, while Local-only training provided an upper bound for client-specific personalization without any knowledge sharing. \texttt{FedYogi} \cite{reddi2020adaptive} implemented adaptive moment-based federated averaging with per-coordinate learning rates. \texttt{FedAdam} \cite{reddi2020adaptive} employed the Adam optimizer \cite{kingma2015adam} with server-side momentum accumulation, \texttt{FedAvgM} \cite{hsu2019measuring} combined momentum acceleration with standard FedAvg updates \cite{mcmahan2017communicationefficient}, \texttt{FedAdagrad} \cite{reddi2020adaptive} utilized Adagrad-style adaptive learning rates, and \texttt{pFedLoRA} \cite{yi2024pfedlora} provided the strongest personalized LoRA baseline in our aligned-setting comparisons. \texttt{PFAdapter} itself follows the sample-count-weighted aggregation rule in \cref{eq:selective_aggregation}, synchronizing only the query/key adapters while keeping the value/output adapters client-private. The base architecture employed \texttt{MiniCPM-V-2\_6-int4}, a quantized multimodal large language model with Qwen2 backbone. LoRA \cite{hu2021lora} was applied to self-attention projection matrices with rank $r=8$ and scaling factor $\alpha_{\text{LoRA}}=16$. Local training utilized \texttt{AdamW} optimizer with learning rate $2 \times 10^{-5}$ and cosine annealing over 50 communication rounds. Each client executed one local epoch with batch size 1 and gradient accumulation steps of 16. Federated training sampled 2 clients per round from a total population of 10. Orthogonality regularization weight was set to $\lambda_{\text{o}} = 0.1$ based on validation performance. Experiments were conducted on a single Nvidia L20 GPU with 48GB VRAM, utilizing 8-bit quantization and gradient checkpointing for memory efficiency.
All tables use this nominal 50-round budget as their common stopping rule. Longer-horizon quantities such as the rounds required to reach a fixed accuracy threshold are treated separately as extended diagnostics in the supplementary material, rather than as alternative stopping points for the main comparisons.

\subsection{Performance Evaluation}
\label{subsec:performance_evaluation}

\noindent\textbf{Main results on aligned modal scenario.}\label{subsec:main_results}
Quantitative comparisons between \texttt{PFAdapter} and state-of-the-art federated learning baselines are presented in \cref{tab:main_results}. Among the learned baselines, \texttt{pFedLoRA} provides the personalized reference across the reported metrics, and \texttt{PFAdapter} improves upon it on every dataset while simultaneously reducing communication overhead by nearly 50\%. On VQA-RAD, \texttt{PFAdapter} attains 62.83\% overall accuracy, exceeding \texttt{pFedLoRA} by 1.09\%. Performance gains remain consistent on SLAKE, where 60.08\% accuracy improves upon the 59.12\% baseline by 0.96\%. For Hateful Memes, \texttt{PFAdapter} reaches 75.23\% accuracy and 75.63\% AUC, surpassing the best baseline scores of 74.38\% accuracy and 74.92\% AUC from \texttt{pFedLoRA}. On CrisisMMD, \texttt{PFAdapter} achieves 62.49\% accuracy, improving upon the baseline result of 61.48\%. Communication analysis further shows that \texttt{PFAdapter} transmits only 315 MB per round compared to 617 MB for the full-adapter baselines, achieving a 48.9\% reduction in bandwidth requirements through selective aggregation of query and key projection modules only.
\Cref{tab:main_results} reports macro-averaged client-local test scores in the aligned setting together with mean $\pm$ standard deviation over three seeds for the learned federated baselines. Larger-client scaling, fairness, and claim-scope details are provided in the supplementary material, \cref{supp:sec:supp_revision_3,supp:sec:supp_revision_4}.
To complement the mean$\pm$std summaries, we further compare \texttt{PFAdapter} against the baseline \texttt{pFedLoRA} using paired two-sided tests over the three aligned-setting seeds, as shown in the supplementary material, \cref{supp:tab:supp_significance_tests}.

\begin{table*}[t]
\centering
\caption{Performance comparison across datasets on the aligned-modal scenario ($\alpha=0.5$). Learned federated baselines are reported as mean $\pm$ standard deviation over three seeds, and best results are highlighted in \textbf{bold}.}
\vspace{-2px}
\label{tab:main_results}
\setlength{\tabcolsep}{4mm}
\resizebox{\linewidth}{!}{
\begin{tabular}{lccccccccc}
\toprule
\rowcolor{gray!8}
\textbf{Method} & \multicolumn{2}{c}{\textbf{VQA-RAD}} & \multicolumn{2}{c}{\textbf{SLAKE}} & \multicolumn{2}{c}{\textbf{Hateful Memes}} & \multicolumn{2}{c}{\textbf{CrisisMMD}} & \textbf{Comm.} \\
\cmidrule(lr){2-3} \cmidrule(lr){4-5} \cmidrule(lr){6-7} \cmidrule(lr){8-9}
\rowcolor{gray!8}
& \textbf{Acc (\%)$\uparrow$} & \textbf{F1 (\%)$\uparrow$} & \textbf{Acc (\%)$\uparrow$} & \textbf{F1 (\%)$\uparrow$} & \textbf{Acc (\%)$\uparrow$} & \textbf{AUC (\%)$\uparrow$} & \textbf{Acc (\%)$\uparrow$} & \textbf{F1 (\%)$\uparrow$} & \textbf{(MB/R)} \\
\midrule
Zero-shot & 56.98 & 52.4 & 64.95 & 61.2 & 66.57 & 65.89 & 24.20 & 22.8 & - \\
Local & 59.64 & 56.8 & 61.63 & 58.9 & 66.39 & 67.12 & 47.34 & 45.2 & - \\
FedYogi \cite{xu2025fedmllm} & $60.53{\pm}0.88$ & $57.9{\pm}0.61$ & $58.67{\pm}0.55$ & $55.4{\pm}0.33$ & $71.41{\pm}0.82$ & $72.48{\pm}1.46$ & $60.82{\pm}0.95$ & $58.6{\pm}0.58$ & 617 \\
FedAdam \cite{reddi2020adaptive} & $60.31{\pm}1.42$ & $57.5{\pm}0.75$ & $56.74{\pm}0.52$ & $53.8{\pm}0.41$ & $72.56{\pm}1.23$ & $73.24{\pm}2.43$ & $59.12{\pm}0.56$ & $57.1{\pm}0.41$ & 617 \\
FedAvgM \cite{hsu2019measuring} & $58.98{\pm}1.35$ & $55.8{\pm}0.67$ & $58.47{\pm}1.06$ & $55.1{\pm}0.47$ & $72.18{\pm}1.71$ & $72.94{\pm}1.61$ & $56.87{\pm}0.42$ & $54.8{\pm}0.53$ & 617 \\
FedAdagrad \cite{reddi2020adaptive} & $60.54{\pm}0.55$ & $57.8{\pm}0.31$ & $55.83{\pm}0.32$ & $52.9{\pm}0.60$ & $73.76{\pm}0.89$ & $73.34{\pm}1.99$ & $60.43{\pm}0.60$ & $58.3{\pm}0.23$ & 617 \\
pFedLoRA \cite{yi2024pfedlora} & $61.74{\pm}0.74$ & $59.1{\pm}0.55$ & $59.12{\pm}0.43$ & $56.8{\pm}0.37$ & $74.38{\pm}0.68$ & $74.92{\pm}0.72$ & $61.48{\pm}0.58$ & $59.4{\pm}0.49$ & 617 \\
\rowcolor{lightkeycolor}
\texttt{PFAdapter} & $\mathbf{62.83{\pm}0.62}$ & $\mathbf{60.1{\pm}0.44}$ & $\mathbf{60.08{\pm}0.38}$ & $\mathbf{57.6{\pm}0.35}$ & $\mathbf{75.23{\pm}0.51}$ & $\mathbf{75.63{\pm}0.63}$ & $\mathbf{62.49{\pm}0.46}$ & $\mathbf{60.3{\pm}0.40}$ & \textbf{315} \\
\bottomrule
\end{tabular}

}
\vspace{-10px}
\end{table*}

\begin{table}[t]
\centering
\caption{
Ablation results reported as accuracy (\%) on VQA-RAD and SLAKE, AUC (\%) on Hateful Memes.
}

\vspace{-2px}
\label{tab:ablation}
\setlength{\tabcolsep}{3mm}
\begin{minipage}{\columnwidth}
\resizebox{\columnwidth}{!}{
\begin{tabular}{c|ccc|cccc}
\toprule
\rowcolor{gray!8}
\textbf{ID} & \textbf{HS} & \textbf{OL} & \textbf{SA} & \textbf{VQA-RAD} & \textbf{SLAKE} & \textbf{Hateful} & \textbf{Comm.} \\
\midrule
1 & \ding{55} & \ding{51} & \ding{51} & 60.53 & 58.67 & 72.48 & 617 \\
2 & \ding{51} & \ding{55} & \ding{51} & 61.33 & 58.58 & 74.30 & 315 \\
3 & \ding{51} & \ding{51} & \ding{55} & 60.93 & 58.18 & 73.80 & 617 \\
\midrule
\rowcolor{lightkeycolor}
4 & \ding{51} & \ding{51} & \ding{51} & \textbf{62.83} & \textbf{60.08} & \textbf{75.63} & \textbf{315} \\
\bottomrule
\end{tabular}
}
\vspace{1px}

{\fontsize{6.5pt}{2pt}\selectfont \quad \textit{Notes}: HS = Hierarchical Split, OL = Orthogonality Loss, SA = Selective Aggregation\par}
\end{minipage}

\vspace{-10px}
\end{table}

\noindent\textbf{Ablation study on component contributions.}\label{subsec:ablation}
To assess the contribution of individual components, we systematically removed each technical module and measured the resulting performance degradation. Removing the orthogonality regularization loss results in accuracy decreases of 1.5\% on VQA-RAD and 1.5\% on SLAKE, together with a 1.33\% AUC decrease on Hateful Memes, validating that knowledge disentanglement between global and local adapters is crucial for effective personalization. Disabling selective aggregation while maintaining orthogonality constraints leads to performance drops of 1.9\%, 1.9\%, and 1.83\% AUC across the three datasets, respectively, while simultaneously doubling communication overhead to 617 MB per round. The largest degradation appears when the hierarchical split is removed, yielding drops of 2.30\% on VQA-RAD, 1.41\% on SLAKE, and 3.15\% AUC on Hateful Memes. That control still uses the same sample-count-weighted server aggregation as \texttt{PFAdapter}. The observed gap therefore isolates the value of the Q/K-global and V/O-local decomposition rather than replacing the server optimizer with \texttt{FedYogi}. Experimental results confirm that: i) explicit disentanglement prevents local adapters from redundantly learning global knowledge, ii) selective aggregation preserves client-specific expertise through private value and output projections, and iii) hierarchical decomposition enables more nuanced control over knowledge sharing than monolithic adapter synchronization.
Detailed weight-washing diagnostics and cross-client attention-map visualizations are provided in the supplementary material, \cref{supp:sec:supp_revision_4}.

\noindent\textbf{Decomposition strategy analysis.}\label{subsec:decomposition} Different module assignment strategies lead to varying performance outcomes depending on which projections are designated for global versus local adaptation. Assigning query ($q_{p}$) and key ($k_{p}$) projections to the global set while keeping value ($v_{p}$) and output ($o_{p}$) projections local yields the optimal configuration, achieving 62.8\% accuracy on VQA-RAD and 60.1\% on SLAKE, as detailed in \cref{tab:decomposition}. Alternative decomposition strategies result in varying degrees of performance degradation. Specifically, assigning $q_{p}$ and $v_{p}$ to global aggregation reduces accuracy by 1.3\% on VQA-RAD, suggesting that value projections are inherently more task-specific and should remain personalized. Restricting global synchronization to only $q_{p}$ leads to a more substantial 2.6\% accuracy decrease, indicating that key projections also capture essential cross-client structural information. Conversely, assigning three modules ($q_{p}, k_{p}, v_{p}$) to the global set reduces communication less substantially and sacrifices 3.1\% accuracy, demonstrating the diminishing returns of excessive global synchronization.
A complementary cross-domain similarity summary in the supplementary material extends this observation beyond VQA-RAD to Hateful Memes, where the query/key branch still exhibits stronger cross-client alignment than the value/output branch. The projection split therefore remains supported by both medical and non-medical evidence rather than by a single-domain visualization alone.
The projection-level evidence directly matches the mechanism in \cref{eq:attention_score,eq:weighted_sum}: removing $k_p$ from the global set degrades cross-client attention alignment, while promoting $v_p$ to the global set erodes the local semantic capacity needed under non-IID supervision \cite{vaswani2017attention,mickus2024role}. The best Q/K-global and V/O-local split therefore emerges not as a heuristic partition, but as the configuration that best preserves relation sharing and client-specific reconstruction simultaneously.

\begin{table}[t]
\centering
\caption{Accuracy (\%) and communication cost (MB/R) of different global-local decomposition strategies on VQA-RAD.}
\vspace{-2px}
\label{tab:decomposition}
\setlength{\tabcolsep}{6.3mm}
\resizebox{\columnwidth}{!}{
\begin{tabular}{llcc}
\toprule
\rowcolor{gray!8}
\textbf{Global} & \textbf{Local} & \textbf{VQA-RAD} & \textbf{Comm.} \\
\rowcolor{gray!8}
\midrule
$q_{p}$ only & $k_{p}, v_{p}, o_{p}$ & 60.2 & 155 \\
$q_{p}, k_{p}, v_{p}$ & $o_{p}$ only & 59.7 & 469 \\
$q_{p}, v_{p}$ & $k_{p}, o_{p}$ & 61.5 & 315 \\
\rowcolor{lightkeycolor}
\textbf{$q_{p}, k_{p}$} & \textbf{$v_{p}, o_{p}$} & \textbf{62.8} & \textbf{315} \\
\bottomrule
\end{tabular}
}
\vspace{-10px}
\end{table}

\begin{figure*}[t]
\centering
\includegraphics[width=\linewidth]{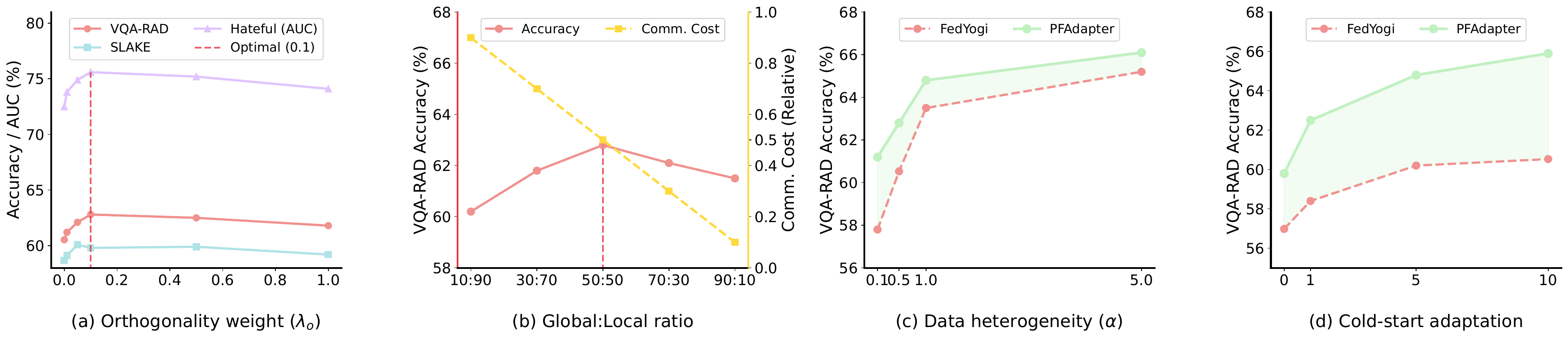}
\caption{Sensitivity analysis of \texttt{PFAdapter}. (a) Impact of orthogonality weight $\lambda_{\text{o}}$. (b) Accuracy vs. communication trade-off for decomposition ratios. (c) Robustness under data heterogeneity (Dirichlet $\alpha$). (d) Cold-start adaptation for new clients.}
\vspace{-2px}
\label{fig:supp_sensitivity}
\vspace{-10px}
\end{figure*}

\noindent\textbf{Parameter sensitivity and robustness analysis.}\label{subsec:sensitivity} Comprehensive sensitivity analysis across three critical hyperparameters reveals optimal configuration ranges and robustness characteristics. \Cref{fig:supp_sensitivity}(a) demonstrates that orthogonality weight $\lambda_{\text{o}} = 0.1$ yields optimal performance for VQA-RAD (62.8\%) and Hateful Memes (75.6\% AUC), while SLAKE achieves peak accuracy at $\lambda_{\text{o}} = 0.05$ (60.1\%). Increasing $\lambda_{\text{o}}$ beyond these optimal values to 0.5 or 1.0 causes gradual degradation, as excessive orthogonality constraints restrict local adapters from capturing client-specific knowledge. \Cref{fig:supp_sensitivity}(b) examines the trade-off between accuracy and communication efficiency across different global-local decomposition ratios, where the 50:50 configuration achieves optimal balance with 62.8\% accuracy at 50\% communication cost. \Cref{fig:supp_sensitivity}(c) evaluates robustness under varying data heterogeneity levels, measured by Dirichlet parameter $\alpha$ ranging from 0.1 (extreme non-IID) to 5.0 (nearly IID). Under high heterogeneity ($\alpha=0.1$), \texttt{PFAdapter} achieves 61.2\% accuracy compared to 57.8\% for \texttt{FedYogi}, representing a 3.4\% improvement that narrows to 0.9\% under low heterogeneity conditions, confirming that hierarchical decomposition provides greater benefits when client distributions diverge more significantly. \Cref{fig:supp_sensitivity}(d) demonstrates cold-start adaptation capability for newly joined edge devices, wherein \texttt{PFAdapter} achieves 59.8\% zero-shot accuracy compared to 56.98\% for \texttt{FedYogi}, and reaches 64.8\% after only 5 local tuning rounds through effective knowledge transfer from pre-aggregated global components.

\begin{figure}[t]
\centering
\includegraphics[width=\linewidth]{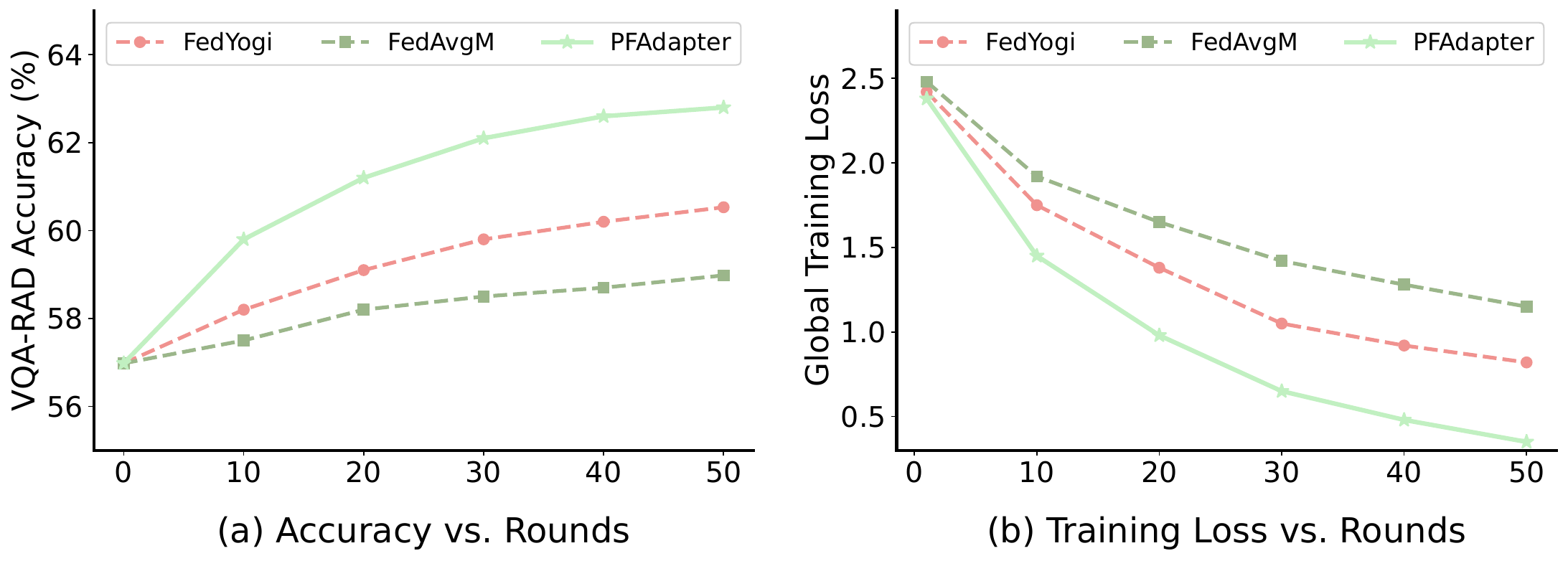}
\caption{Convergence analysis on VQA-RAD. (Left) Accuracy vs. communication rounds. (Right) Training loss reduction.}
\vspace{-2px}
\label{fig:convergence}
\vspace{-10px}
\end{figure}

\noindent\textbf{Convergence behavior and training dynamics.}\label{subsec:convergence} Orthogonality-driven decomposition accelerates training convergence by reducing parameter conflicts during federated aggregation. \texttt{PFAdapter} exhibits significantly faster convergence over 50 communication rounds on VQA-RAD, reaching 61.2\% accuracy by round 20 compared to 59.1\% for \texttt{FedYogi} and 58.2\% for \texttt{FedAvgM}, as illustrated in \cref{fig:convergence}(a).
Faster convergence can be attributed to the orthogonality constraint, which reduces parameter conflicts between local and global updates and facilitates more stable aggregation. In particular, the gradients in \cref{eq:grad_global_ortho,eq:grad_local_ortho} penalize overlap between the global and local update subspaces, so the server aggregates less mutually contradictory information from different clients at each round. \cref{fig:convergence}(b) demonstrates that \texttt{PFAdapter} achieves consistently lower training loss throughout the optimization process, with final loss of 0.35 compared to 0.82 for \texttt{FedYogi} and 1.15 for \texttt{FedAvgM}, indicating a better-optimized loss landscape and more efficient utilization of the parameter budget.

\noindent\textbf{Computational and communication efficiency.}\label{subsec:efficiency} Selective aggregation substantially reduces network resource requirements while maintaining computational efficiency comparable to baseline LoRA methods. Compared to full model fine-tuning requiring 45.2 minutes per round and 42.5 GB peak VRAM, LoRA-based methods reduce training time by approximately 75\% and memory usage by 63\%, as summarized in \cref{tab:efficiency}.
\texttt{PFAdapter} incurs a marginal 5\% increase in training time (10.8 vs. 10.3 min/round for \texttt{FedYogi}) due to the additional orthogonality loss computation, but achieves a 48.9\% reduction in total communication volume over the common 50-round budget (15.75 GB vs. 30.85 GB for \texttt{FedYogi}). Communication savings are achieved by transmitting only the global-shared query and key projection adapters (2 of 4 LoRA modules), while preserving client-specific value and output projections locally. The supplementary target-reaching table reports a separate extended diagnostic in which the methods continue beyond the nominal 50 rounds until 60\% VQA-RAD accuracy is reached, so those rounds-to-target numbers should be read as longer-horizon measurements rather than as alternative main-table stopping budgets.
Peak VRAM consumption of 15.8 GB remains comparable to baseline LoRA methods, making \texttt{PFAdapter} suitable for deployment on single-GPU systems without requiring specialized distributed computing infrastructure. At the same time, the present efficiency evidence remains a controlled communication-and-runtime study rather than a full end-to-end straggler/dropout emulation. Communication payload, heterogeneity stress tests, and client-scaling diagnostics are therefore complemented with an explicit deployment-limitation discussion in \cref{subsec:discussion} and in the supplementary material.

\begin{table}[t]
\centering

\caption{Comparison of train time (min/round), GPU-hours per round, peak VRAM (GB), inference time, and total communication (GB) efficiency.}
\vspace{-2px}
\label{tab:efficiency}
\setlength{\tabcolsep}{1mm}
\resizebox{\columnwidth}{!}{\begin{tabular}{lccccc}
\toprule
\rowcolor{gray!8}
\textbf{Method} & \textbf{Train Time} & \textbf{GPU-h/R} & \textbf{Inf. Time} & \textbf{Peak VRAM} & \textbf{Total Comm.} \\
\rowcolor{gray!8}
\midrule
Full Tuning & 45.2 & 0.753 & 125.4 & 42.5 & 125.4 \\
FedYogi \cite{xu2025fedmllm} & 10.3 & 0.172 & 14.7 & 15.7 & 30.85 \\
FedAvgM \cite{hsu2019measuring} & 10.6 & 0.177 & 14.2 & 15.5 & 30.85 \\
\rowcolor{lightkeycolor}
\texttt{PFAdapter} & \textbf{10.8} & \textbf{0.180} & \textbf{14.8} & \textbf{15.8} & \textbf{15.75} \\
\bottomrule
\end{tabular}
}
\vspace{-10px}
\end{table}

\begin{figure*}[t]
\centering
\includegraphics[width=\textwidth]{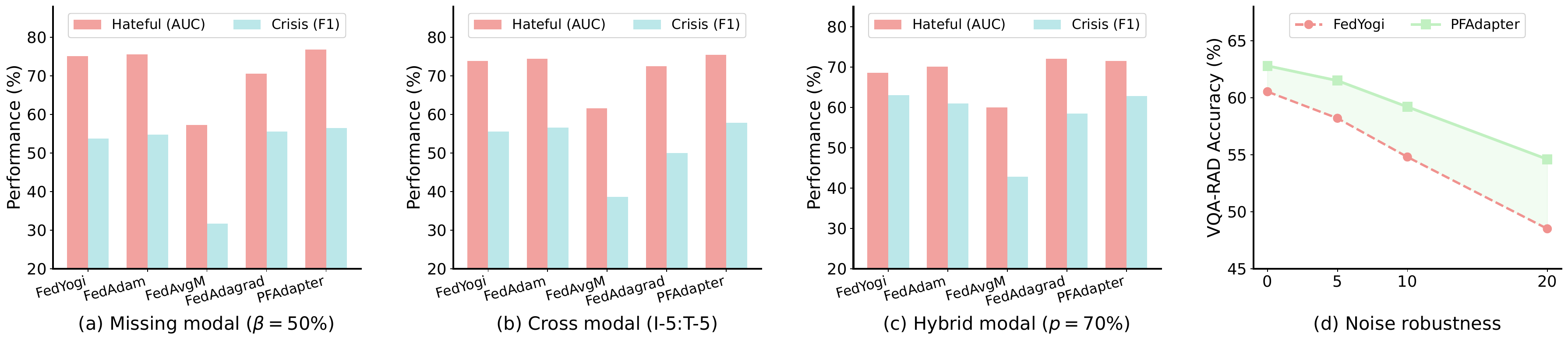}
\caption{Robustness analysis across heterogeneity scenarios, including Missing, Cross, Hybrid modalities, and noise levels.}
\vspace{-2px}
\label{fig:robustness}
\vspace{-10px}
\end{figure*}

\noindent\textbf{Robustness across multimodal heterogeneity scenarios.}~Selective parameter aggregation enables \texttt{PFAdapter} to maintain consistent performance advantages across diverse challenging scenarios. In the Missing Modal scenario where 50\% of clients lack either image or text modalities, \texttt{PFAdapter} achieves 76.8\% AUC on Hateful Memes and 56.5\% F1 on CrisisMMD, outperforming FedYogi by 1.68\% and 2.72\%, respectively, as visualized in \cref{fig:robustness}(a). Subsequently, \cref{fig:robustness}(b) evaluates generalization when image-dominant and text-dominant clients coexist (I-5:T-5 split), with \texttt{PFAdapter} maintaining superior performance across all three metrics. Moreover, \cref{fig:robustness}(c) combines aligned ($p=70\%$) and missing modal conditions, where the local adaptation capability of $v_p$ and $o_p$ parameters proves particularly beneficial. Finally, \cref{fig:robustness}(d) demonstrates graceful degradation under increasing Gaussian noise levels, with \texttt{PFAdapter} maintaining 54.6\% accuracy at 20\% noise compared to 48.5\% for FedYogi, achieving a 6.1\% absolute advantage. Collectively, experimental results validate that selective aggregation and hierarchical decomposition provide inherent robustness to diverse real-world data distribution challenges.

\vspace{-5px}
\subsection{Discussion and Limitations}\label{subsec:discussion}
Superior performance of \texttt{PFAdapter} stems from three interrelated factors: i) \textit{Structural decomposition} recognizes functional heterogeneity within self-attention mechanisms, wherein query and key projections encode cross-attention patterns amenable to global sharing across the network, whereas value and output projections modulate edge-specific representations, ii) \textit{Orthogonality regularization} prevents redundant learning by enforcing local adapters to capture orthogonal directions in parameter space relative to global knowledge, leading to more efficient parameter budget utilization across distributed agents, and iii) \textit{Selective aggregation} reduces the weight washing effect common in federated MLLM fine-tuning, wherein client-specific adaptations become diluted through uniform parameter mixing. 
Nevertheless, several limitations merit discussion regarding deployment in heterogeneous edge networks: i) Optimal decomposition ratios may vary depending on the degree of local-global divergence across network nodes, with 50:50 split providing the best balance for moderate heterogeneity ($\alpha=0.5$) yet potentially requiring adaptation for extreme distribution shifts, ii) Current framework applies uniform decomposition across all transformer layers, whereas layer-specific ratios based on sensitivity analysis could further enhance performance for edge devices with varying computational capabilities, iii) The present systems evidence covers synchronized payload, runtime, and controlled client/data heterogeneity, but does not yet include a dedicated distributed testbed with explicit straggler, upload-cap, or client-dropout scheduling, and iv) Although orthogonality regularization effectively disentangles knowledge, Frobenius norm constraints may not fully capture complex non-linear dependencies between global and local parameters in highly dynamic network environments. Future work could explore learnable decomposition ratios, layer-adaptive splitting strategies tailored to network topology, asynchronous or bandwidth-aware federation protocols, and more sophisticated disentanglement metrics based on information-theoretic measures suitable for agentic AI systems.

\section{Conclusion}
\label{sec:conclusion}

In this paper, we presented \texttt{PFAdapter}, a communication-efficient personalized federated learning framework for deploying multimodal large language models as intelligent agents across heterogeneous edge networks. Hierarchical LoRA decomposition was introduced to explicitly separate adapter parameters into global-shared and local-private components based on functional roles of self-attention modules. Query and key projections are assigned to global synchronization across the federated network, whereas value and output projections remain localized for edge-specific adaptation. Orthogonality regularization enforces effective knowledge disentanglement between network-synchronized and edge-retained parameters. Selective aggregation protocols transmit only global-shared components, reducing communication overhead by nearly 50\% while preserving edge-specific expertise.
Future research directions include exploring dynamic decomposition ratios adapted to network conditions, layer-adaptive splitting strategies for heterogeneous edge devices, and integration with emerging 6G network to further enhance agentic AI deployment across next-generation communication systems.

\begin{spacing}{0.95}
\bibliographystyle{IEEEtran}
\bibliography{refs}
\end{spacing}

\end{document}